\documentclass{article}

\usepackage[utf8]{inputenc}
\usepackage{listings}
\usepackage{xcolor}
\usepackage[linesnumbered,ruled,vlined]{algorithm2e}

\usepackage[round]{natbib}
\usepackage{graphicx}
\usepackage{amsmath}
\usepackage{amsfonts}
\usepackage{amssymb}
\usepackage[affil-it]{authblk} 
\usepackage{comment}

\usepackage{hyperref}
\hypersetup{
	colorlinks=true,
	linkcolor=blue,
	filecolor=magenta,      
	urlcolor=cyan,
}

\title{Hybrid Ranking Network for Text-to-SQL}
\author{Qin Lyu, Kaushik Chakrabarti, Shobhit Hathi, Souvik Kundu, Jianwen Zhang, Zheng Chen}
\affil{Microsoft Dynamics 365 AI\\\normalsize{\{qilyu, kaushik, shhathi, sokun, jiazhan, zhengc\}@microsoft.com}}

\begin{document}
	\date{}
	\maketitle
	
	\begin{abstract}
		In this paper, we study how to leverage pre-trained language models in Text-to-SQL. We argue that previous approaches under utilize the base language models by concatenating all columns together with the NL question and feeding them into the base language model in the encoding stage. We propose a neat approach called Hybrid Ranking Network~(HydraNet) which breaks down the problem into column-wise ranking and decoding and finally assembles the column-wise outputs into a SQL query by straightforward rules. In this approach, the encoder is given a NL question and one individual column, which perfectly aligns with the original tasks BERT/RoBERTa is trained on, and hence we avoid any ad-hoc pooling or additional encoding layers which are necessary in prior approaches. Experiments on the WikiSQL dataset show that the proposed approach is very effective, achieving the top place on the leaderboard.
	\end{abstract}
	
	\section{Introduction}
	\label{intro}
	Relational databases are prevalent in many real-world applications. Typically, a structured query language such as SQL is required to interact with such databases. However, mastering SQL queries is generally difficult. It has been a long standing goal to enable users interacting with relational databases via human natural languages. The general problem was known as ``Natural Language Interface to Databases~(NLIDBs)'' in database areas~\citep{androutsopoulos1995natural,popescu2003towards,Affolter2019}. In recent years, there has been a surge of interest on deep learning-based approaches to a crucial problem of NLIDBs: translating a natural language query to SQL, often referenced as ``NL-to-SQL'' or ``Text-to-SQL''~\citep{zhongSeq2SQL2017,xu2017sqlnet, yu2018naacl, shi2018incsql, hwang2019sqlova, he2019xsql, guo2019irnet}. 
	
	In this paper, we focus on the Text-to-SQL problem and experiment with the WikiSQL~\footnote{\url{https://github.com/salesforce/WikiSQL}}~ dataset~\citep{zhongSeq2SQL2017}. WikiSQL is the first large-scale dataset for Text-to-SQL, with about 80K human annotated pairs of NL question and SQL query. WikiSQL constrained the problem by two factors -- each question is only addressed by a single table, and the table is known. This constrained setting has guided research to focus on the core elementary problem. Even though the scope is constrained, the dataset is still very challenging because the tables and questions are very diverse. Notably, there are about 24K different tables associated with this dataset.
	
	In the past, several approaches were proposed for the WikiSQL dataset. They primarily share the similar encoder-decoder architecture. The encoder encodes information from both the NL question and the table schema into some hidden representation. Some of them encode an NL question with the full table schema, e.g., concatenating the NL question with all the column names~\citep{zhongSeq2SQL2017,dong2018acl,chang2019zero,hwang2019sqlova,he2019xsql}, while others encode an NL question with each column separately~\citep{xu2017sqlnet,yu2018naacl,hwang2019sqlova}. There are also some work do both at different layers~\citep{hwang2019sqlova,he2019xsql}. The decoder decodes the hidden representation to a SQL query. Some early work tried the ``to sequence'' style one step decoding~\citep{zhongSeq2SQL2017,dong2018acl} however it is found challenging to guarantee the output correct in syntax. Later more work breaks down a SQL query to several parts like SELECT column, WHERE column, WHERE operator, WHERE value, etc. and have sub-decoders for them~\citep{xu2017sqlnet,yu2018naacl,hwang2019sqlova,he2019xsql}. In this way, the chance of output violating syntax is reduced. Beyond improving the Text-to-SQL task models, recently there are some work study how to leverage pre-trained language models~\citep{devlin2018bert,liu2019roberta,yang2019xlnet,liu2019multi} and get promising results~\citep{hwang2019sqlova,he2019xsql}. All the previous work reveal several major challenges in Text-to-SQL on WikiSQL: (1) How to fuse information from both NL question and table schema, which is handled by encoder; (2) How to make sure the output SQL query executable and accurate, which is handled by decoder; (3) How to leverage pre-trained language models.
	
	This paper is primarily motivated by the above challenge (3), however, the proposed approach contributes to solving all the above challenges. We argue that the previous approaches~\citep{hwang2019sqlova,he2019xsql} did not align the task model well with the base language model, and hence, the power of the base language model is compromised by the task model. Specifically, both \citet{hwang2019sqlova} and \citet{he2019xsql} concatenate the full table schema with the NL question and feed into a BERT-alike base language model. However, in their decoding stage, both need the hidden representation for each individual column. Thus, they have to apply adhoc pooling on the tokens of a column name to get a vector for the column. \citet{hwang2019sqlova} simply pick the vector for the first token in their ``shallow layer'' and added another LSTM layers in their ``decoder-layer'' and ``NL2SQ-Layer''. \citet{he2019xsql} apply weighted average over vectors of all tokens. These ad-hoc pooling operations and additional LSTM layers caused information loss and bring in unnecessary complexity. To solve the dilemma encountered in previous work, in this paper, we choose to encode a pair of NL question and one individual column via the exact BERT/RoBERTa model structure. Then we have multiple sub-tasks in decoder stage: SELECT \& WHERE column ranking, condition operator, and condition value span. Since the decoder does not output final SQL query, we apply straightforward rules to assemble the decoder outputs to a final SQL query. Our approach has two benefits. First, the inputs, i.e., a question and column pair align perfectly with the original sentence pair training tasks of BERT/RoBERTa, and hence, we believe the power of the BERT/RoBERTa is utilized best. Second, as we only encode one column, the ``[CLS]'' vector in BERT/RoBERTa output captures all fused information from the question and the column, which is exactly the ``column vector'' needed for the decoder. So we don't need to apply any further pooling or additional complex layers. This makes our model structure very simple and efficient. Since the main philosophy in our approach is ranking on columns but with multi-task outputs, we name it Hybrid Ranking Network (HydraNet).
	
	In summary, our contributions are in three folds. First, we propose a simple and efficient network structure which perfectly aligns the Text-to-SQL task with the base language models, and hence, the power of the base language models are best utilized. Second, the base language model as encoder directly encodes an NL question and a column without any additional pooling operations which is believed to be the best encoder capturing the question-column relation in Text-to-SQL. Third, the proposed hybrid ranking mechanism and execution-guided decoding handle column-column relations and effectively boost the accuracy. In other words, our approach resolved all the aforementioned challenges at the same time. Our approach achieves the best result on WikiSQL dataset, which verifies its effectiveness and contributions.

	\section{Related Work}	
	Previous work on WikiSQL roughly fall into the following three categories. 
	\begin{itemize}
		\item \textbf{Sequence-to-sequence translation}: Early works on this task follow this approach~\citep{zhongSeq2SQL2017,dong2018acl}. They do not work well since the order among the WHERE conditions or those among the items in SELECT clause do not matter in SQL. The order, however, matters for the sequence-to-sequence model and finding the best ordering is typically hard. One approach to mitigate the ordering issue is to employ reinforcement learning \citep{zhongSeq2SQL2017}. However, the improvement is typically limited (e.g., 2\% improvement reported for WikiSQL task by \citep{zhongSeq2SQL2017}).
		
		\item \textbf{Sub-tasks for different parts of an SQL query}:  The next set of approaches break down an SQL query into different parts, such as \texttt{SELECT\-column}, \texttt{SELECT-aggregation}, \texttt{WHERE-number}, \texttt{WHERE-column}, \texttt{WHERE-\\operator},
		\texttt{WHERE-value}, etc. Following, each part is handled by dedicated sub-models \citep{xu2017sqlnet,yu2018naacl}. These approaches avoid the ordering issue but many of the interdependencies are not leveraged, such as the dependence between \texttt{SELECT-column} and \texttt{WHERE-column}, or the dependence between \texttt{WHERE-columns} of two WHERE conditions. In addition, they generally use pre-trained word representations like GloVe but unfortunately no contextualized word representations like BERT or MT-DNN are used (as they were developed before BERT and MT-DNN). Our approach belongs to this category but we utilized BERT/RoBERTa.
		
		\item \textbf {Approaches leveraging pre-trained language models}: The most recent approaches~\citep{hwang2019sqlova, he2019xsql} leverage pre-trained language models like BERT and MT-DNN which achieved state-of-the-art performance on many NLP tasks. Usually, application tasks would build on top of these language models and do fine tuning with the task specific data. However, it is not a trivial problem to design the task model structure so that the power of base language models can be best utilized. Both ~\citet{hwang2019sqlova} and \citet{he2019xsql} construct the input to the encoding layer by concatenating all candidate columns together with the NL question. The benefit is they can capture the interdependencies between \texttt{SELECT-column} and \texttt{WHERE-column} as well as \texttt{WHERE-column}s in different WHERE conditions. The drawback is that they finally need vectors for individual columns, and they have to apply pooling or additional LSTM encoding layer to form a column vector. In the decoding stage, they used the aforementioned way of defining sub-tasks for different parts of SQL query. In contrast, in the encoder, our input is a pair of an NL question and an individual column. As we pointed out in Sec.~\ref{intro}, our approach is more coherent with the base language models and we do not need any further pooling or additional encoding layers.
	\end{itemize}

	Similar to us, \citet{xu2017sqlnet} also formulates the problem as column-wise prediction. However, there are two major differences: first, it does not use pre-trained language models, and second, they separately train multiple models for sub-tasks while we use a single model for all the sub-tasks. \citet{yu2018naacl} extended the approach of \citet{xu2017sqlnet} by including the {\em entity type} information into the question. Different to them, we include the {\em data type} information of column in our approach. On the question side, we argue that it is not necessary since it is believed that the common entity types can be captured by the base language models.

	\section{Approach}
	In this section, we formulate Text-to-SQL as a multi-task learning problem which can be solved by adapting a pre-trained Transformer model. We describe our model's expected input representation, our method of transforming SQL queries into this representation, and the training tasks which produce the SQL output. We then describe our column ranking method, our model's training and inference, and finally our execution guided decoding method.   
	
	\subsection{Input Representation}
	Given a question $q$ and candidate columns $c_1, c_2, \dots, c_k$, we form one pair of input sentences for each column as $(\text{Concat}(\phi_{c_i}, t_{c_i}, c_i), q)$, where $\phi_{c_i}$ is the type of column $c_i$ in text (e.g., ``string'', ``real'', etc.), $t_{c_i}$ is the table name of the table that $c_i$ belongs to, and $\text{Concat}(\cdot)$ is a function to concatenate a list of texts into one string. In this paper, we simply join the texts with blank space.
	
	The input sentences pair will be further tokenized and encoded to form the inputs of Transformer-based model. For example, if we choose BERT as the base model, the token sequence will be:
	
	$$[\text{CLS}], x_1, x_2, ..., x_{m}, [\text{SEP}], y_{1}, y_{2}, ..., y_{n}, [\text{SEP}]$$
	where $x_1, x_2, \dots, x_{m}$ is the token sequence of column representation $\text{Concat}(\phi_{c_i}, t_{c_i}, c_i)$, and $y_{1}, y_{2}, ..., y_{n}$ is the token sequence of question $q$. The token sequences will be encoded by encoder of the base model to form the final inputs of model.
	
	\subsection{SQL Query Representation and Tasks}
	\label{sqlrep}
	In this paper, we consider SQL queries with no nested structure~\footnote{HydraNet can be extended to support more complex form of SQL queries, including nested queries. Details will be coming soon.}, and having the following form:
	\begin{lstlisting}
	"sql":{
	"select": [(agg1, scol1), (agg2, scol2), ...]
	"from": [table1, table2, ...]
	"where":[(wcol1, op1, value1), (wcol2, op2, value2), ...]
	}
	\end{lstlisting}
	
	We divide objects in above SQL query into two categories:
	
	\begin{enumerate}
		\item Objects associated with specific column, e.g., aggregation operator, value text span. 
		\item Global objects with no association with specific column, e.g., select\_num (number of select clauses) and where\_num (number of where conditions).
	\end{enumerate}
	
	For each column-question input pair $(c_i, q)$, the prediction of objects in 1 can be formulated as sentence pair classifications or question answering tasks. More specifically, we denote the output sequence embedding of base model by $h_{[\text{CLS}]}, h^{c_i}_{1}, \dots, h^{c_i}_{m}, h_{[\text{SEP}]}, h^{q}_{1}, \dots, h^{q}_{n}, h_{[\text{SEP}]}$,
	
	\begin{enumerate}
		\item For an aggregation operator $a_j$, let $P(a_j|c_i, q) = \text{softmax}(W^{agg}[j, :] \cdot h_{[\text{CLS}]})$. During training, we mask out columns that are not in a select clause for aggregation operator training task. 
		
		\item For a condition operator $o_j$, let $P(o_j|c_i, q) = \text{softmax}(W^{op}[j, :] \cdot h_{[\text{CLS}]})$. During training, we mask out columns that are not in a where clause for condition operator training task. 
		
		\item For value start and end indices, let $P(y_j=\text{start}|c_i, q) = \text{softmax}(W^{start}\cdot h^{q}_{j})$ and $P(y_j=\text{end}|c_i, q) = \text{softmax}(W^{end}\cdot h^{q}_{j})$. During training, the start and end index is set to 0 for columns that are not in where clause. This aligns with the no answer setting for BERT in question answer task \citep{devlin2018bert}.
	\end{enumerate}
	
	For global objects with no association, note that $P(z|q) = \sum_{c_i}P(z|c_i, q)P(c_i|q)$. We formulate the calculation of $P(z|c_i, q)$ as sentence pair classification, and $P(c_i|q)$ as the similarity between column $c_i$ and question $q$. Specifically,
	
	\begin{enumerate}
		\item For number of select clauses $n_s$, let $P(n_s|q) = \sum_{c_i}P(n_s|c_i, q)P(c_i|q)$
		\item For number of where clauses $n_w$, let $P(n_w|q) = \sum_{c_i}P(n_w|c_i, q)P(c_i|q)$
	\end{enumerate}
	The calculation of $P(c_i|q)$ will be presented in the next section.
	
	\subsection{Column Ranking}
	\label{colrank}
	For each question $q$, we denote $\mathcal{S}_q$ to be the set of columns that are in SELECT clause, and $\mathcal{W}_q$ as the set of columns that are in WHERE clause. Let $\mathcal{R}_q \dot= \mathcal{S}_q\cup\mathcal{W}_q$ be the set of columns that are in the SQL query, i.e., relevant column set. Finally, we denote the candidate column set as $\mathcal{C}_q=\{c_1, c_2, \dots, c_k\}$. Obviously, $\mathcal{R}_q\subseteq\mathcal{C}_q$.
	
	There are three ranking tasks:
	\begin{enumerate}
		\item SELECT-Rank: Rank $c_i\in\mathcal{C}_q$ based on whether the SELECT clause of $q$ contains $c_i$, i.e., whether $c_i\in \mathcal{S}_q$. 
		\item WHERE-Rank: Rank $c_i\in\mathcal{C}_q$ based on whether the WHERE clause of $q$ contains $c_i$, i.e., whether $c_i\in \mathcal{W}_q$.  
		\item Relevance-Rank: Rank $c_i\in\mathcal{C}_q$ based on whether the SQL query of $q$ contains $c_i$, i.e., whether $c_i\in \mathcal{R}_q$.  
	\end{enumerate}
	
	BERT is a strong interaction-based matching model and has demonstrated strong effectiveness on ranking tasks\citep{qiao2019understanding}. It is recommended in \citep{qiao2019understanding} to use $w\cdot h_{[\text{CLS}]}$ as ranking score, and fine-tune base model using classification loss. Hence, we let $P(c_i\in \mathcal{S}_q |q) = \text{sigmoid}(w_{sc}\cdot h_{[\text{CLS}]})$, $P(c_i\in \mathcal{W}_q |q) = \text{sigmoid}(w_{wc}\cdot h_{[\text{CLS}]})$ and $P(c_i\in \mathcal{R}_q |q) = \text{sigmoid}(w_{rc}\cdot h_{[\text{CLS}]})$ be the ranking scores for SELECT-Rank, WHERE-Rank and Relevance-Rank task, respectively. 
	
	With SELECT-Rank scores $P(c_i\in \mathcal{S}_q |q)$, we can rank and choose the top candidate columns to form SELECT clause. To determine the number of columns to keep, we can either 1. set a threshold and only keep those columns that have scores above the threshold, or 2. directly set the number to the predicted select\_num proposed in Section \ref{sqlrep}, i.e.,
	\begin{equation}
	\label{eqn:select_num}
	\hat{n}_s=\arg\max_{n_s}P(n_s|q)=\arg\max_{n_s}\sum_{c_i\in\mathcal{C}_q} P(n_s|c_i, q)P(c_i\in \mathcal{R}_q |q)
	\end{equation}
	
	In this paper, we choose the second approach and keep the top $\hat{n}_s$ columns to form SELECT clause.
	
	Similarly, for WHERE clause, we rank candidate columns by $P(c_i\in \mathcal{W}_q |q)$, and choose the top $\hat{n}_w$ columns to form WHERE clause, where
	\begin{equation}
	\label{eqn:where_num}
	\hat{n}_w=\arg\max_{n_w}P(n_w|q)=\arg\max_{n_w}\sum_{c_i\in\mathcal{C}_q} P(n_w|c_i, q)P(c_i\in \mathcal{R}_q |q)
	\end{equation}

	\subsection{Training and Inference}
	In training phase, each labeled sample $(q_i, \mathcal{R}_{q_i}), \mathcal{C}_{q_i}=\{c_{q_i1}, c_{q_i2}, \dots, c_{q_i n_i}\}$ is first transformed into $n_i$ column-question samples $(c_{q_i1}, q_i) ,(c_{q_i2}, q_i), \dots,  (c_{q_i n_i}, q_i)$. Then the SQL query label of $(q_i, \mathcal{C}_{q_i})$ is parsed and used to label each of the column-question samples for all the individual tasks in Section \ref{sqlrep} and \ref{colrank}. The optimization object is the summation of the cross-entropy loss for each of the individual tasks, over all the column-question samples generated from original samples $(q_1, \mathcal{C}_{q_1}), (q_2, \mathcal{C}_{q_2}), \dots, (q_n, \mathcal{C}_{q_n})$. Note that we shuffle the column-question samples during training, hence each batch contains samples from different questions and columns.
	
	During inference, we first get predicted class labels for each of the individual tasks from model output. Then the predicted SQL query is constructed in the following steps:
	
	\begin{enumerate}
		\item Get predicted select\_num and where\_num via equation \ref{eqn:select_num} and \ref{eqn:where_num}, respectively.
		\item Rank $c_i\in\mathcal{C}_q$ by predicted SELECT-Rank scores, and keep the top $\hat{n}_s$ columns $\hat{sc}_1, \hat{sc}_2 \dots, \hat{sc}_{\hat{n}_s}$. The SELECT clause is then set to
		$$[(\hat{agg}_1, \hat{sc}_1), (\hat{agg}_2, \hat{sc}_2), \dots, (\hat{agg}_{\hat{n}_s}, \hat{sc}_{\hat{n}_s})]$$ where $\hat{agg}_i$ is the predicted aggregation operator of $\hat{sc}_i, i=1,2,\dots, \hat{n}_s$.
		
		\item Rank $c_i\in\mathcal{C}_q$ by predicted WHERE-Rank scores, and keep the top $\hat{n}_w$ columns $\hat{wc}_1, \hat{wc}_2 \dots, \hat{wc}_{\hat{n}_w}$. The WHERE clause is then set to $$[(\hat{wc}_1, \hat{op}_1, \hat{val}_1), (\hat{wc}_2, \hat{op}_2, \hat{val}_2), \dots, (\hat{wc}_{\hat{n}_w}, \hat{op}_{\hat{n}_w}, \hat{val}_{\hat{n}_w})]$$
		where $\hat{op}_i$, $\hat{val}_i$ is the predicted condition operator and predicted value text of column $\hat{wc}_i, i=1,2,\dots, \hat{n}_w$.
		
		\item Denote by $\hat{\mathcal{T}}=\{\hat{t}_1, \hat{t}_2, \dots, \hat{t}_{n_t}\}$ the union set of tables of columns $\hat{sc}_1, \hat{sc}_2 \dots, \hat{sc}_{\hat{n}_s}$ and $\hat{wc}_1, \hat{wc}_2 \dots, \hat{wc}_{\hat{n}_w}$. The FROM clause is set to $$[\hat{t}_1, \hat{t}_2, \dots, \hat{t}_{n_t}]$$
	\end{enumerate}
	
	\subsection{Execution-guided decoding}
	\label{sec:eg}
	Neural network models predict SQL query based on syntactic and semantic information extracted from input query, and from column-value relation learnt from training data. However, these could still be not enough for making good prediction at run-time, as
	\begin{enumerate}
		\item Database values and columns have discrete relation and can change from time to time without constraints. A trained model could be missing the latest database information and be misguided by the out-dated column-value relation captured in model weights.
		\item Model outputs for each task is predicted independently, and could generate invalid combinations. For example, a string-type column is not allowed to combine with an aggregation operator, or condition operator like greater-than. Although a model could learn from training data to assign low probabilities to such combinations, it is still not guaranteed to eliminate such cases.
	\end{enumerate}
	
	To address these issues, \citet{2018executionguided} proposed Execution-guided decoding (EG), which executes predicted SQL query at run-time and makes correction if database engine returned run-time error or empty output. Based on their idea, we apply execution-guided decoding after model prediction, which is described in Algorithm \ref{alg:EG}.
	
	\begin{algorithm}
		\label{alg:EG}
		\SetAlgoLined
		\KwResult{Executable SQL query}
		Set execution SQL query $Q_e$ to empty. Set predicted SQL query $Q_p$ to empty\;
		Get top-$k_1$ predicted aggregation-SELECT pairs $(\hat{agg}_1, \hat{sc}_1), \dots, (\hat{agg}_{k_1}, \hat{sc}_{k_1})$ via beam search (Note that the predicted columns $\hat{sc}_1, \dots, \hat{sc}_{k_1}$ could be duplicated)\;
		\For{i=1 to $k_1$}{
			\If{$\hat{sc}_i$ not in $Q_p[\text{``select"}]$}
			{
				Set $Q_e[\text{``select"}]=[(\hat{agg}_i, \hat{sc}_i)]$\;
				result = Execute($Q_e$)\;
				\If{result is not empty}
				{$Q_p[\text{``select"}]$.Add($(\hat{agg}_i, \hat{sc}_i)$)}
			}
			\If{$len(Q_p[\text{``select"}])=\hat{n}_s$}
			{break\;}
		}
		Get top-$k_2$ predicted conditions $(\hat{wc}_1, \hat{op}_1, \hat{val}_1), \dots, (\hat{wc}_{k_2}, \hat{op}_{k_2}, \hat{val}_{k_2})$ via beam search (Note that the predicted columns $\hat{wc}_1, \dots, \hat{wc}_{k_2}$ could be duplicated)\;
		\For{i=1 to $k_2$}{
			\If{$\hat{wc}_i$ not in $Q_p[\text{``where"}]$}
			{
				Set $Q_e[\text{``select"}]=[(\text{null}, \hat{wc}_i)]$ and $Q_e[\text{``where"}]=[(\hat{wc}_i, \hat{op}_i, \hat{val}_i)]$\;
				result = Execute($Q_e$)\;
				\If{result is not empty}
				{$Q_p[\text{``where"}]$.Add($(\hat{wc}_i, \hat{op}_i, \hat{val}_i)$)}
			}
			\If{$len(Q_p[\text{``where"}])=\hat{n}_w$}
			{break\;}
		}
		return $Q_p$\;
		
		\caption{Execution-guided Decoding for HydraNet}
	\end{algorithm}
	
	\section{Experiment}
	In this section, we demonstrate results of HydraNet on the WikiSQL dataset and compare it to the other state-of-the-art approaches.
	
	WikiSQL dataset \citep{zhongSeq2SQL2017} contains 56,355, 8,421, and 15,878 question-SQL query pairs for training, development and testing, respectively. All the SQL queries have exactly one select column and aggregation operator, and 0 to 4 conditions. A table is given for each query, hence we set candidate columns to be the columns in the given table.
	
	Table \ref{tab:wikisql} shows results of different approaches, both with and without applying execution-guided decoding (EG). Logical form accuracy is the percentage of exact matches of predicted SQL queries and labels, and execution accuracy is the percentage of exact matches of executed results of predicted SQL queries and labels. We use logical form accuracy on development dataset as metric to choose the best model. The results show that on test set, HydraNet is consistently better than the other approaches. Note that HydraNet with BERT-Large-Uncased is significant better than SQLova, which uses the same base model, and is even as good as X-SQL, which uses MT-DNN as base model. MT-DNN has shown to be significantly better than BERT-Large-Uncased~\citep{liu2019multi}, and has similar score as RoBERTa on GLUE Benchmark~\footnote{https://gluebenchmark.com/leaderboard}. This implies that HydraNet is better at exploiting the pre-trained Transformer model.
	
	By comparing accuracy on development and test set, we find that HydraNet also shows better generalization, which we believe is because it only adds dense layers to the output of base model, which is simpler and has less parameters than the output architectures of X-SQL and SQLova.
	
	\begin{table}[h!]
		\centering
		\begin{tabular}{ |c|c||c|c| }
			\hline
			Model & Base Model & Dev (lf, ex) & Test (lf, ex) \\\hline
			\hline
			SQLova & BERT-Large-Uncased & 81.6, 87.2 & 80.7, 86.2 \\ 
			X-SQL & MT-DNN & 83.8, 89.5 & 83.3, 88.7 \\ 
			HydraNet & BERT-Large-Uncased & 83.5, 88.9 & 83.4, 88.6\\
			HydraNet & RoBERTa-Large & 83.6, 89.1 & \textbf{83.8}, \textbf{89.2} \\\hline
			
			SQLova + EG & BERT-Large-Uncased & 84.2, 90.2 & 83.6, 89.6 \\ 
			X-SQL + EG & MT-DNN & 86.2, 92.3 & 86.0, 91.8 \\ 
			HydraNet + EG & BERT-Large-Uncased & 86.6, 92.2 & 86.2, 91.8\\
			HydraNet + EG & RoBERTa-Large & 86.6, 92.4 & \textbf{86.5}, \textbf{92.2} \\
			
			\hline
		\end{tabular}
		\caption{Logical form (lf) and execution (ex) accuracy on WikiSQL dataset}
		\label{tab:wikisql}
	\end{table}
	
	Table \ref{tab:wikisql2} shows the accuracy of each task with and without applying EG. On test set, HydraNet has the best SELECT column accuracy, and almost the same WHERE column accuracy as X-SQL. This proves that the column-question pair ranking mechanism is as good as question-with-all-column approach in comparing columns and selecting the relevant ones. After applying EG, we find that HydraNet achieved almost equal accuracy as X-SQL on tasks relate to condition prediction, i.e., W-COL, W-OP and W-VAL. This proves that execution-guided decoding is also necessary in capturing column-column relations, especially in helping model to correct condition value predictions. In sum, column-question input is as sufficient as question-with-all-column input, as long as ranking mechanism and execution-guided decoding is applied to capture column-column relations.
	
	\begin{table}[t]
		\centering
		\resizebox{\columnwidth}{!}{
			\begin{tabular}{ |c||c|c|c|c|c|c| }
				\hline
				Model & S-COL & S-AGG & W-NUM & W-COL & W-OP & W-VAL \\\hline
				\hline
				SQLova & 97.3, 96.8 & 90.5, 90.6 & 98.7, 98.5 & 94.7, 94.3 & 97.5, 97.3 & 95.9, 95.4 \\ 
				X-SQL &  97.5, 97.2 & 90.9, 91.1 & 99.0, \textbf{98.6} & 96.1, \textbf{95.4} & 98.0, \textbf{97.6} & 97.0, \textbf{96.6} \\
				HydraNet & 97.8, \textbf{97.6} & 91.1, \textbf{91.4} & 98.7, 98.4 & 95.7, 95.3 & 97.8, 97.4 & 96.2, 96.1 \\\hline
				SQLova + EG & 97.3, 96.5 & 90.7, 90.4 & 97.7, 97.0& 96.0, 95.5& 96.4, 95.8& 96.6, 95.9 \\ 
				X-SQL+ EG &  97.5, 97.2 & 90.9, 91.1& 99.0, \textbf{98.6} & 97.7, \textbf{97.2} & 98.0, \textbf{97.5} & 98.4, \textbf{97.9} \\
				HydraNet + EG & 97.8, \textbf{97.6} & 91.1, \textbf{91.4} & 98.8, 98.4 & 97.5, \textbf{97.2} & 97.9, \textbf{97.5} & 98.1, 97.6  \\\hline
		\end{tabular}}
		\caption{The dev and test accuracy of each task on WikiSQL dataset. S-COL, S-AGG, W-NUM, W-COL, W-OP and W-VAL stands for tasks of predicting SELECT column, aggregation operator, number of conditions, WHERE columns, WHERE operators and WHERE values, respectively.}
		\label{tab:wikisql2}
	\end{table}
	
	\section{Conclusion}
	In this paper, we study how to leverage pre-trained language models like BERT on WikiSQL task. We formulate text-to-SQL as a column-wise hybrid ranking problem and propose a neat network structure called HydraNet which best utilizes the pre-trained language models. The proposed model has a simple architecture but achieves No.1 result on the WikiSQL leaderboard. We believe it will bring in deep insight to the Text-to-SQL problem on how to better utilize pre-trained language models. In the future, we will extend the capability of HydraNet to support full SQL grammar.
	
	\bibliographystyle{plainnat}
	\bibliography{references}
\end{document}